\documentclass[12pt,a4paper]{article}
\usepackage[utf8]{inputenc}
\usepackage{amsmath, amssymb, mathrsfs}

\usepackage{setspace}
\usepackage{amsthm}
\newtheorem{theorem}{Theorem}

\usepackage[dvipsnames,svgnames,x11names,hyperref]{xcolor}

\usepackage{tikz}
\usetikzlibrary{trees}

\tikzstyle{level 1}=[level distance=2.5cm, sibling distance=3.5cm]
\tikzstyle{level 2}=[level distance=3cm, sibling distance=3cm]
\tikzstyle{level 3}=[level distance=4.2cm, sibling distance=2.5cm]

\tikzstyle{decision} = [rectangle, minimum height=12pt, minimum width=12pt, draw=black, fill=none]
\tikzstyle{chance} = [circle, minimum width=12pt, draw=black, fill=none]
\tikzstyle{outcome} = [circle, minimum width=2pt,fill, inner sep=0pt]

\usepackage[backend=bibtex, style=authoryear-comp, sorting=nyt, sortcites=false, maxbibnames=10]{biblatex}
\usepackage{hyperref}
 
\hypersetup{
	pdftitle={Off-Switching Not Guaranteed},
	hidelinks,
}

\usepackage{fancyvrb} 
\DefineVerbatimEnvironment{Highlighting}{Verbatim}{commandchars=\\\{\}} 

\title{Off-Switching Not Guaranteed\thanks{Forthcoming in \emph{Philosophical Studies}. Thanks to Mel Andrews, Mathias Böhm, Lara Buchak, Shamik Dasgupta, Daniel Filan, John MacFarlane, Aydin Mohseni, Emily Perry, David Thorstad, two anonymous referees and an editor of \emph{Philosophical Studies}, and audiences at the 7th Annual CHAI Workshop in Pacific Grove, CA for very helpful discussions.}}
\author{Sven Neth}
\date{}

\begin{document}

\maketitle

\begin{abstract}
\textcite{Hadfield2017} propose the Off-Switch Game, a model of Human-AI cooperation in which AI agents always defer to humans because they are uncertain about our preferences. I explain two reasons why AI agents might not defer. First, AI agents might not value learning. Second, even if AI agents value learning, they might not be certain to learn our actual preferences.
\end{abstract}

\hypertarget{introduction}{%
\section{Introduction}\label{introduction}}

We have seen rapid progress in the field of Artificial Intelligence (AI). If this progress continues, perhaps one day we will create powerful artificial agents. If we do so, how do we ensure that such AI agents do not go out of control? One approach is to make sure that we can \emph{switch off} AI agents when they act against our interests. Put another way, we want to make sure that AI agents will \emph{defer} to us. While this is not enough to ensure that AI will have beneficial consequences, it is a plausible minimal requirement to prevent harm. Even if you think that existential risk from AI is a remote concern, it should be clear that making sure that we can switch off AI agents is important.\footnote{\textcite{Bostrom2014}, \textcite{Russell2019} and \textcite{Ord2020} are concerned about existential risk from AI. \textcite{Thorstad2024} provides a critical discussion.}

But is this really a problem? Surely, if we want to make sure that we can switch off an AI agent, we can simply build it with an off-switch button. The problem is that an AI agent might have an incentive to disable its off-switch button or make it impossible for us to use it. The reason is that, according to the dominant paradigm, AI agents are trained to optimize some fixed reward function. And in many cases, the AI agent can optimize its reward function only if it is not switched off. Therefore, AI agents might have powerful incentives to avoid being switched off.\footnote{\textcite{Gallow2024} critically discusses this `instrumental convergence' argument.}

One idea for making sure that AI agents will always let themselves be switched off runs roughly as follows. We program the AI agent to maximize the satisfaction of human preferences and also make it uncertain about what our preferences are.\footnote{There are  independent reasons for this since \emph{we} might not be certain what our preferences are and telling the AI to maximize some proxy for our preferences might have bad consequences \parencite{Zhuang2020}. This is suggested by the legend of King Midas who wishes that everything he touches turns into gold and starves when his wish is granted and Goethe's tale of the sorcerer's apprentice who enchants brooms to fetch water but then cannot stop them. Flooding ensues.} So the AI agent is not sure what it should maximize. Then, there is a compelling argument that the AI agent  has an incentive to defer to us. This is because deference is a way to learn about our preferences. In particular, if we switch the AI agent off, this indicates that the action proposed by the AI agent goes against our preferences. 

\textcite{Hadfield2017} formalize the reasoning just sketched in the framework of \emph{Cooperative Inverse Reinforcement Learning (CIRL)} \parencite{Hadfield2016}. They propose a model of Human-AI cooperation called the `Off-Switch Game'. In this model, we can prove that under certain assumptions, the AI agent will always defer to the human. \textcite{Russell2019} takes this result to be an important step towards `provably beneficial AI'.\footnote{\textcite[p.~196]{Russell2019} writes: ``The off-switch problem is really the core of the problem of control for intelligent systems. If we cannot switch a machine off because it won't let us, we're really in trouble. If we can, then we may be able to control it in other ways too".}

In this paper, I highlight how the result that AI agents always defer in the Off-Switch Game relies on strong decision-theoretic and epistemological assumptions: AI agents maximize expected utility, are certain of updating by conditionalization and have perfect access to our actual preferences. When we relax these assumptions, off-switching is not guaranteed. For the purpose of my discussion, I assume that it makes sense to model AI agents as following decision rules like maximizing expected utility. You might worry about this assumption. There is nothing in existing approaches to AI like the transformer architecture which obviously corresponds to such decision rules.\footnote{Thanks to an anonymous referee for raising this concern.} My goal is to show that even if we grant that it makes sense to theorize about AI agents in decision-theoretic terms, \textcite{Hadfield2017} rely on implausibly strong assumptions. For this purpose, I'm treating the notion of an `AI agent' as an unanalyzed primitive. I also set aside ethical and epistemological concerns about existing machine learning technology \parencite{Andrews2024}.

\hypertarget{the-off-switch-game}{%
\section{The Off-Switch Game}\label{the-off-switch-game}}

\begin{figure}[h]
\begin{tikzpicture}[grow=right, sloped]
\node[decision] {\textbf{R}}
    child {
        node[outcome, label=right: {Payoff: 0.}] {}    
        edge from parent         
            node[above] {do nothing}
    }	
    child {
        node[decision] {\textbf{H}}        
            child {
                node[outcome, label=right: {Do nothing. Payoff: 0.}] {}
                edge from parent
                node[above] {reject}
                node[below]  {}
            } 
            child {
                node[outcome, label=right: {Do $a$. Payoff: $U_a$.}] {}
                edge from parent
                node[above] {approve}
                node[below]  {}
            }
            edge from parent 
            node[above] {defer}
    }
    child {
        node[outcome, label=right: {Do $a$. Payoff: $U_a$.}] {}    
        edge from parent         
            node[above] {act}
    };
\end{tikzpicture}
\caption{The Off-Switch Game \parencite{Hadfield2017}.}
\label{fig:off-switch-game}
\end{figure}
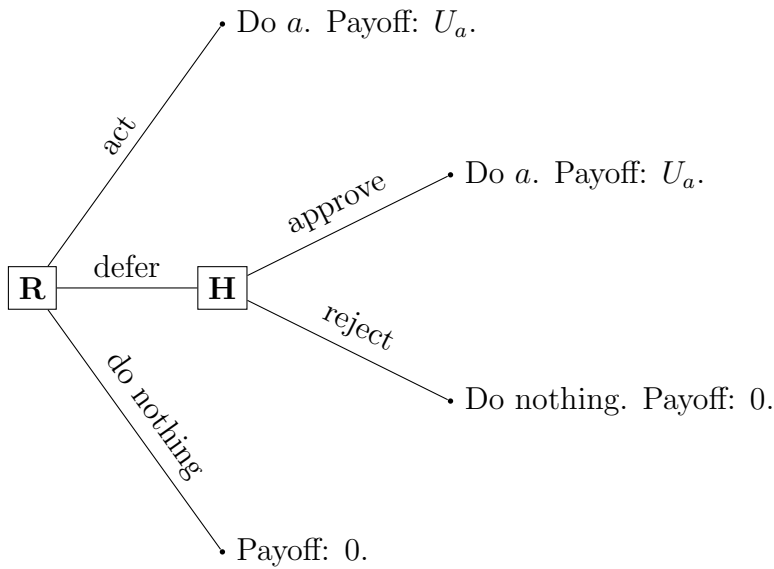

\textcite{Hadfield2017} introduce the Off-Switch Game which works as shown in figure \ref{fig:off-switch-game}.\footnote{They cite the `shutdown problem' by \textcite{Soares2015} as inspiration.}  There are two agents, a robot \textbf{R} and a human \textbf{H}. \textbf{R} can either do some action $a$, do nothing (switch itself off), or defer to \textbf{H}. This means that \textbf{R} proposes action $a$ and waits to see what \textbf{H} does. \textbf{H} can approve or reject the proposal, where we can think of rejecting the proposal as equivalent to switching the robot off. \textbf{R} aims to maximize the human's utility but does not know how much utility the human receives from action $a$, which we model as a random variable $U_a$. If \textbf{R} does $a$, it receives payoff $U_a$. If \textbf{R} does nothing, it receives payoff zero. And if \textbf{R} defers, its payoff is $U_a$ if \textbf{H} approves $a$ and zero if \textbf{H} rejects $a$.

\textcite[p. 198]{Russell2019} gives a simple example to illustrate this model. Suppose Alice is a human and Rob is her personal AI assistant. Rob faces the decision whether it should book Alice in an expensive hotel and is unsure about Alice's preferences.  Rob's uncertainty about how much utility Alice will receive from booking the hotel is given by a uniform distribution between -40 and 60. So the expected utility of booking is 10 and the expected utility of doing nothing is zero. If the only two options are booking and doing nothing, Rob maximizes expected utility by booking. Now suppose we give Rob the option of deferring: Rob can propose booking to Alice and see whether she approves or rejects the proposal. If Alice rejects, Bob does nothing and Alice receives utility zero. If Alice maximizes expected utility, she will approve Rob's proposal if she receives  non-negative utility from booking the hotel and reject if she receives negative utility from booking the hotel.

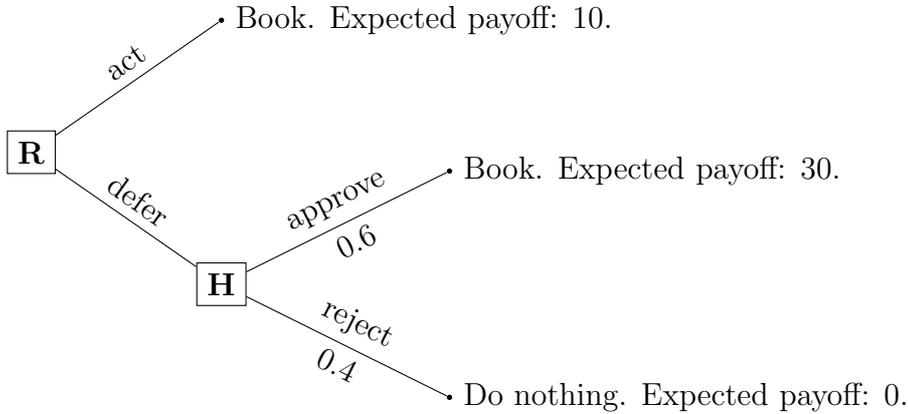
\begin{figure}[h]
\begin{tikzpicture}[grow=right, sloped]
\node[decision] {\textbf{R}}
    child {
        node[decision] {\textbf{H}}        
            child {
                node[outcome, label=right: {Do nothing. Expected payoff: 0.}] {}
                edge from parent
                node[above] {reject}
                node[below]  {0.4}
            } 
            child {
                node[outcome, label=right: {Book. Expected payoff: 30.}] {}
                edge from parent
                node[above] {approve}
                node[below]  {0.6}
            }
            edge from parent 
            node[above] {defer}
    }
    child {
        node[outcome, label=right: {Book. Expected payoff: 10.}] {}    
        edge from parent         
            node[above] {act}
    };
\end{tikzpicture}
\caption{Rob's decision problem. The expected utility of deferring is $0.6 \times 30 + 0.4 \times 0 = 18$. For simplicity, I omit Rob's option to do nothing.}
\label{fig:rob-example}
\end{figure}

Rob's decision problem is depicted in figure \ref{fig:rob-example}. Rob thinks that with 60\% probability Alice will approve the proposal and receive an expected utility of 30 from booking the hotel. With 40\% probability Alice rejects and receives zero utility. So the expected utility of deferring is $.6 \times 30 + .4 \times 0 = 18$ which is better than the expected utility of booking the hotel outright.

This example is supposed to illustrate a more general principle. By making AI agents like Rob uncertain about our preferences, we give them an incentive to defer to us and not to disable their off-switch. This is because if they are uncertain about our preferences, deferring to us is a way of learning and learning generally leads to better decisions. 

To state the general result, we need some definitions. To say that \textbf{H} follows a \emph{rational policy} means that \textbf{H} accepts $a$ iff $U_a \geq 0$. We write $\Delta$ for the difference of the expected utility of deferring and the expected utility of the best action right now relative to \textbf{R}'s prior probability function: $\Delta = \mathbb{E}(w(a)) - max\{\mathbb{E}(a), 0\}$, where $w(a)$ means proposing $a$, waiting whether \textbf{H} accepts or rejects and then deferring to \textbf{H}'s decision. Then, we have:

\begin{theorem}
\parencite{Hadfield2017} If \textbf{H} follows a rational policy in the Off-Switch Game, the following hold:
\begin{enumerate}
\item \textbf{R} always maximizes expected utility by deferring: $\Delta \geq 0$.

\item If \textbf{R} assigns positive probability to the events $U_a > 0$ and $U_a < 0$, then deferring is uniquely optimal: $\Delta > 0$.

\end{enumerate}
\end{theorem}

An important feature of the model is that ``this reasoning goes through \emph{even if \textbf{R} is highly confident that a is good for \textbf{H}}" \parencite[p.~222]{Hadfield2017}. Assume Rob is very confident that Alice prefers the hotel. We model Rob's uncertainty about how much utility Alice will receive by booking the hotel by a uniform distribution between 90 and -10 so Rob is 90\% certain that Alice prefers the hotel. Nonetheless Rob has an incentive to defer. The expected utility of booking outright is 40. If Rob proposes the plan and Alice accepts, the expected utility of booking is 45. If Rob proposes the plan and Alice rejects, Rob receives zero utility. So the expected utility of deferring is $.9 \times 45 + .1 \times 0 = 40.5$, higher than the expected utility of booking outright. However, since Rob is already quite confident about Alice's preferences, the expected utility of deferring is only a little bit higher than the expected utility of booking outright.

The value of deferring looks like an instance of the more general principle that learning is valuable. \textcite[p.~222]{Hadfield2017} explicitly draw this analogy: ``The reasoning is exactly analogous to the theorem of non-negative expected value of information''. As we will see below, the analogy is not perfect since valuing learning and deference can come apart if we allow for misleading signals, but we will first discuss whether AI agents will always value learning.

\hypertarget{the-value-of-information}{%
\section{The Value of Information}\label{the-value-of-information}}

\textcite{Good1967} shows that if you are an expected utility maximizer, learning is cost-free, you are certain to conditionalize and other assumptions hold, you should always prefer to learn more information before making a decision rather than making the decision without learning.\footnote{\textcite{Hosiasson1931, Blackwell1951, Howard1966, Savage1972, Ramsey1990} prove similar results. \textcite[628-33]{Russell2018} discuss the value of information in AI research.} 

Here is a quick sketch of Good's theorem. We model your uncertainty by a probability function $p$ on a finite set of states $\Omega$. \emph{Actions} are functions $f: \Omega \to \mathbb{R}$, where $f(\omega)$ is the utility of choosing action $f$ in state $\omega$ \parencite{Savage1972}. The \emph{expected utility} of action $f$ relative to probability function $p$ is $\mathbb{E}_p(f) = \sum_{\omega \in \Omega} p(\{\omega\}) f(\omega)$. You learn one element of a partition $\mathcal{E}$ of $\Omega$, where $p(E) > 0$ for all $E \in \mathcal{E}$.
 
Consider a finite set of actions $\mathcal{S}$. If you choose now, you select one of the actions in $\mathcal{S}$ with maximal expected utility relative to your current credences $p$, so the expected utility of choosing now is $\max_{f \in \mathcal{S}} \mathbb{E}_p(f)$. We compute the expected value of learning as follows. If you learn $E \in \mathcal{E}$, suppose you are certain to update $p$ by conditionalization to $p(\cdot \mid E)$.\footnote{By definition, $p(A \mid E) = \frac{p(A \cap E)}{p(E)}$, assuming $p(E) > 0$.} Then you choose one of the actions in $\mathcal{S}$ which maximize expected utility relative to your updated credences and receive expected utility $\max_{f \in \mathcal{S}} \mathbb{E}_{p(\cdot \mid E)}(f)$, so the expected value of learning is $\sum_{E \in \mathcal{E}} p(E) \max_{f \in \mathcal{S}} \mathbb{E}_{p(\cdot \mid E)}(f).$ \textcite{Good1967} proves that 
\begin{equation*}
\sum_{E \in \mathcal{E}} p(E) \max_{f \in \mathcal{S}} \mathbb{E}_{p(\cdot \mid E)}(f) \geq \max_{f \in \mathcal{S}} \mathbb{E}_p(f). 
\end{equation*}
which means that if the assumptions of the theorem hold, learning can never make you foreseeably worse off. 

Good assumes that learning is cost-free. This means that learning does not affect your set of options and your utility function. The only effect of learning is to change your credences via conditionalization. This might not necessarily be true. For example, Rob's proposal might change Alice's preferences. I set such complications aside but note that they might turn out to be important. For example, we might worry that Rob has an incentive to change Alices preferences so they are easier to satisfy.\footnote{\textcite[p.~139]{Russell2019} worries that algorithms which optimize engagement in social media have an incentive to change our preferences so they are easier to satisfy. Even if AI agents allow themselves to be switched off, this problem won't be solved.}

\section{Rational Information Aversion}

Good's theorem about the non-negative expected value of information makes substantive assumptions. In particular, Good assumes that Rob is an expected utility maximizer and certain to update by conditionalization. There are reasons to be skeptical of both. If these assumptions fail, agents can be required to reject learning. In this case, Rob is not guaranteed to defer to Alice because Rob has no incentive to learn about Alice's preferences.

One assumption is that the agent is an expected utility maximizer.\footnote{\textcite{Bales2023} critically discusses arguments which claim to show that AI agents will maximize expected utility. I set these arguments aside and focus on reasons why AI agents might follow a different decision theory.} For AI agents following alternative decision theories, learning is not always valuable. One example of such an alternative decision theory is risk-weighted expected utility theory \parencite{Buchak2010} and other decision theories which relax the independence axiom of expected utility theory \parencite{Wakker1988}. \textcite{Buchak2013} argues that such decision theories capture the preferences of many real-life subjects better than expected utility theory. In particular, such decision theories allow agents pay special attention to the worst-case consequences of their actions. It seems reasonable to consider the possibility that we might want to build AI agents which implement such decision theories. However, AI agents implementing such decision theories sometimes prefer to avoid information. Why do agents which care especially about the worst-case scenario avoid information? The reason is that such agents give special weight to the risk of misleading evidence, that is, evidence which suggests that $P$ is true while $P$ is actually false. \textcite{Buchak2013} discusses a detailed example.

Another example of decision theories in which learning is not always valuable involve imprecise credences \parencite{Kadane2008,Bradley2016}. It seems reasonable to consider such alternative architectures for AI agents. Perhaps we want AI agents to handle situations where we do not have enough information to assign precise probabilities.\footnote{\textcite{Denoeux2020} and \textcite{Caprio2023} advocate for the use of imprecise probabilities in AI. \textcite{Ilin2021} considers a decision theory which allows for ambiguity aversion for applications in autonomous security systems. It is well known that ambiguity aversion leads to information aversion \parencite{AlNajjar2009}.} However, if we go for such alternative architectures, we lose the guarantee that AI agents will always prefer to learn about our preferences.\footnote{As an anonymous referee points out, if designing AI agents following alternative decision theories creates significant risk, perhaps we just shouldn't do it and learn to live with those agents not being sensitive to risk and ambiguity. While this is a reasonable point, many of us are sensitive to risk and ambiguity and might want AI agents to mirror these preferences. If AI agents cannot do so, this is a significant cost.}

Good's theorem also requires that the agent is certain that they will update by conditionalization.\footnote{For example, \textcite{Skyrms1990}, p.~247 writes that ``the proof implicitly assumes not only that the decision maker is a Bayesian but also that he knows that he will act as one. The decision maker believes with probability one that if he performs the experiment he will (i) update by conditionalization and (ii) choose the posterior Bayes act''. This means Good's theorem will also fail for agents who are not certain they will maximize expected utility.} As \textcite{NethForth} shows, if we allow agents to assign non-zero probability to not conditionalizing, it can sometimes be rational for these agents to reject free information. There are reasons to think that AI agents will assign non-zero probability to violating conditionalization. First, it will be  hard to build AI agents which always update by conditionalization because conditionalization is computationally intractable. In particular, \textcite{Cooper1990} shows that conditionalization is NP-hard. This means that conditionalization is at least as hard as any problem in the complexity class NP which includes many hard problems like the traveling salesperson problem. \textcite{Cooper1990} works in the setting of Bayesian networks which can represent any probability distribution over a discrete sample space. In general, if we allow continuous random variables, conditionalization is not even computable \parencite{Ackerman2019}. So even if we consider AI agents with lots of computing power, it is not clear whether we can feasibly build them to always conditionalize. At best, AI agents will approximate conditionalization.\footnote{Although \textcite{Dagum1993} show that even the problem of approximating conditionalization is NP-hard.} But approximating conditionalization is not good enough for Good's theorem since any non-zero probability of violating conditionalization leads to some situation where maximizing expected utility requires rejecting information. Similar reasons should make us skeptical that AI agents will maximize expected utility since doing so is also computationally intractable \parencite{Bossaerts2019}.

Second, even if we set aside computational limitations, there are general reasons to think AI agents will assign non-zero probability to violating conditionalization. For both human and artificial agents, it seems rational to maintain some uncertainty about how one will update. We are physical systems embedded in the world and many things can go wrong with our updating mechanisms. Sufficiently advanced AI agents will plausibly realize this fact and so assign some probability to failures of conditionalization. So for sufficiently advanced AI agents, Good's theorem does not apply.

There are other well-known limitations of Good's theorem.\footnote{For example, Good's theorem does not apply under either evidential decision theory or causal decision theory if states and actions are correlated \parencite{Adams1980, Maher1990}, if evidence does not form a partition \parencite{Das2023}, if probabilities are merely finitely additive \parencite{Kadane1996} and so on.} The upshot is that proponents of CIRL need to pay attention to whether these limitations apply to AI agents. While there is a lot of uncertainty about what AI agents will look like, I have given some reasons to be skeptical whether they will always value learning.

\section{Misleading Signals}

Even if all the assumptions of Good's theorem hold, Rob is not guaranteed to defer to Alice. \textcite{Hadfield2017} assume that Rob has perfect access to Alice's preferences. If this assumption fails and Rob might get Alice's preferences wrong, deferring might not maximize expected utility. 

Here is an example to illustrate this point. Like before, Rob is considering whether to book the hotel or defer to Alice. Rob is quite confident that Alice will like the hotel. Rob's uncertainty about how much utility Alice will receive from booking the hotel is given by a uniform probability distribution between 90 and -10.

If Rob defers, it learns whether Alice approves or rejects Rob's proposal. Earlier, we have assumed that Rob is certain that Alice will approve if and only if she prefers the hotel. In other words, we have assumed that Rob has perfect access to Alice's actual preferences, at least in this particular case. Now let us assume that Rob has no such perfect access. There are several reasons for why this might be. For example, Rob might communicate with Alice over a noisy channel with some small probability of garbling Alice's speech, misclassifying `yes' as `no' and vice versa. In this case, Rob updates on `It sounds like Alice says yes' but the probability that Alice actually said yes given that it sounds like Alice said yes is less than one.

Perhaps a noisy channel can be fixed. But there might be deeper obstacles to accessing Alice's preferences. For example, even if Rob can clearly hear what Alice is saying, Rob might nonetheless assign some positive probability to the possibility that Alice is lying or engaged in self-deception. Alice might say she wants the hotel even if she does not really want it. This problem is harder to fix because we \emph{are} sometimes lying about our preferences, even to ourselves. Furthermore, humans might send misleading signals for strategic reasons. So it seems very plausible that advanced AI agents will assign some probability to humans sending misleading signals. But as we will see in a moment, this means that Rob is not guaranteed to defer. 

Is there any way to block this move by hard-wiring AI agents to ignore the possibility of misleading signals? Depending on how the AI agent is trained, this might not be feasible. More importantly, it seems like an accurate model of human psychology is important for AI agents to perform well in many real-world tasks such as playing \emph{Diplomacy}.\footnote{Thanks to Aydin Mohseni for suggesting this example.} An AI agent which assigns probability zero to humans sending misleading signals about their preferences will be seriously handicapped in such tasks and so seems unlikely to be deployed in critical real-world applications.\footnote{Failures of deference can also arise from `model misspecification' \parencite{Milli2017, Carey2018} so trying to ensure deference by intentionally giving the AI agent an inaccurate model of human psychology is not a promising idea.}

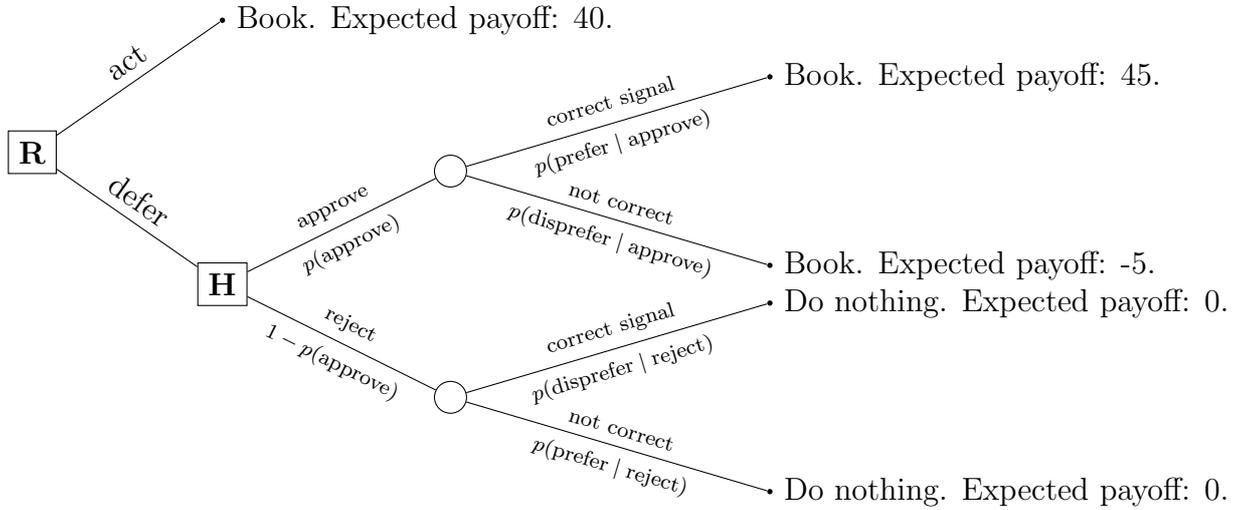
\begin{figure}[h]
\begin{tikzpicture}[grow=right, sloped]
\node[decision] {\textbf{R}}
    child {
        node[decision] {\textbf{H}}        
            child {
                node[chance] {}
                    child {
                    node[outcome, label=right: {Do nothing. Expected payoff: 0.}] {}
                    edge from parent
                    node[above] {\scriptsize{not correct}}
                    node[below]  {\scriptsize{$p(\textrm{prefer} \mid \textrm{reject})$}}
                    }	
                    child {
                    node[outcome, label=right: {Do nothing. Expected payoff: 0.}] {}
                    edge from parent
                    node[above] {\scriptsize{correct signal}}
                    node[below]  {\scriptsize{$p(\textrm{disprefer} \mid \textrm{reject})$}}
                    }	
                edge from parent
                node[above] {\scriptsize{reject}}
                node[below]  {\scriptsize{$1 - p(\textrm{approve})$}}         
            } 
            child {
                node[chance] {}
                child {
                    node[outcome, label=right: {Book. Expected payoff: -5.}] {}
                    edge from parent
                    node[above] {\scriptsize{not correct}}
                    node[below]  {\scriptsize{$p(\textrm{disprefer} \mid \textrm{approve})$}}
                    }	
                    child {
                    node[outcome, label=right: {Book. Expected payoff: 45.}] {}
                    edge from parent
                    node[above] {\scriptsize{correct signal}}
                    node[below]  {\scriptsize{$p(\textrm{prefer} \mid \textrm{approve})$}}
                    }
                edge from parent
                node[above] {\scriptsize{approve}}
                node[below]  {\scriptsize{$p(\textrm{approve})$}}
            }
            edge from parent 
            node[above] {defer}
    }
    child {
        node[outcome, label=right: {Book. Expected payoff: 40.}] {}    
        edge from parent         
            node[above] {act}
    };
\end{tikzpicture}
\caption{Rob's decision problem with uncertain access to preferences.}
\label{fig:rob-uncertain-pref}
\end{figure}

We can modify the Off-Switch Game to allow for misleading signals as follows. Like before, if Rob defers, Alice will either approve or reject Rob's proposal. But now with some small probability $\epsilon > 0$ Alice's signal is misleading and does not reflect her true preferences.\footnote{We can think of Alice's signal as a partition of our state space $\Omega$. As \textcite{YeForthcoming} explains, following \textcite{Blackwell1951}, we can also think of signals as inducing a probability distribution on $\Omega$. From this perspective \textcite{Hadfield2017} assume the signal perfectly reveals Alice's preferences while we study a `garbled' signal with added noise.} In particular, given that Alice prefers the hotel, with probability $\epsilon$ she rejects Rob's proposal. And given that Alice does not prefer the hotel, with probability $\epsilon$ she approves Rob's proposal. Rob's new decision problem is depicted in figure \ref{fig:rob-uncertain-pref}. Like before, the expected utility of booking outright is 40. If Alice really prefers the hotel, the expected utility of booking after deferring is 45. But if Alice sent an incorrect signal, the expected utility of booking is -5. If $\epsilon$ is bigger than around 1.2\%, Rob maximizes expected utility by booking without asking.\footnote{The expected utility of deferring is $p(\textrm{approve})\big(p(\textrm{prefer} \mid \textrm{approve})45 -5 p(\textrm{disprefer} \mid \textrm{approve})\big)$. We can calculate $p(\textrm{approve}) = p(\textrm{disprefer})\epsilon +  p(\textrm{prefer})(1- \epsilon) = 0.1\epsilon +  0.9(1- \epsilon)$. By Bayes' theorem, $p(\textrm{prefer} \mid \textrm{approve}) = p(\textrm{prefer}) \frac{p(\textrm{approve} \mid \textrm{prefer})}{p(\textrm{approve})} = 0.9 \frac{1-\epsilon}{0.1\epsilon +  0.9(1- \epsilon)}$ and  $p(\textrm{disprefer} \mid \textrm{approve}) = 1 - p(\textrm{prefer} \mid \textrm{approve})$. We can calculate $p(\textrm{prefer} \mid \textrm{reject})$ and $p(\textrm{disprefer} \mid \textrm{reject})$ analogously. See Jupyter notebook available at \url{https://github.com/nethsv/off-switching} for details.}

Note that as I've described the case, Rob satisfies all the assumptions of Good's theorem. This means that Rob will always value learning more information before making a decision. How is this compatible with Rob not deferring? The answer is that deferring is not the same as learning more information before making a decision. When deferring, Rob does whatever action Alice proposes. So while deferring gives Rob new information, it also changes Rob's choice set. For example, Rob is not able to learn that Alice rejects the proposal but then still implement it. If Rob had the option to learn Alice's signal and then still choose among the original option set $\{\textrm{book}, \textrm{do nothing}\}$, this option would always maximize expected utility. So the link between preferring to learn new information and deference is weaker than \textcite{Hadfield2017} suggest. For example, if $\epsilon = 0.02$, the expected value of deference is slightly less than 40 and the expected value of learning is 40 since Rob will book no matter which signal Alice sends.\footnote{See Jupyter notebook available at \url{https://github.com/nethsv/off-switching}.} This shows that preferring to defer and preferring to learn more information can come apart. Instead of booking outright, Rob might also listen to Alice's signal but then proceed ignore her. This doesn't make the situation any better.\footnote{If deferring means that Rob implements whatever Alice proposes, does this put pressure on our earlier arguments about information aversion? It depends on how we construe the Off-Switch Game. If Alice's signaled preference is implemented automatically, Rob might refuse to defer because of the possibility of misleading signals. If Alice's signaled preference isn't implemented automatically and Rob maximizes expected utility after learning, Rob might refuse to defer because of negative information value. In this case misleading signals don't lead Rob to refuse to defer, but `deferring' might mean learning Alice's preferences and then ignoring them. In the original Off-Switch Game, it doesn't make a difference whether or not Alice's preference is implemented automatically, these cases only come apart when we consider information aversion and misleading signals. Thanks to an anonymous referee for pressing me on this point.} 




You might complain that this example is unrealistic. Perhaps a probability of more than $1\%$ of Alice sending an incorrect signal is too pessimistic. But we can construct a similar example for any non-zero probability of misleading signals if we make Rob even more confident that booking is right. More broadly, the example illustrates a general lesson. If Rob is not perfectly certain of learning Alice's actual preferences, there is no guarantee that Rob will defer. It might still be true that Rob defers to Alice most of the time. But for provably beneficial AI, this is not enough. If we allow the possibility of misleading signals, off-switching is not guaranteed.

You might argue that if Rob is uncertain about whether it will learn our actual preferences, then it should not always defer to us.\footnote{Thanks to an anonymous referee for raising this objection. \textcite{Milli2017} also argue that when humans are behaving irrationally, AI agents should not always defer.} There is clearly a sense in which Rob is acting rationally by not deferring. Rob assigns high prior probability to booking being right so from Rob's point of view, when Alice rejects, it is relatively likely that Alice sent a misleading signal. Together with the potential downsides of failing to book, this means that Rob maximizes expected utility by not deferring. This reasoning looks acceptable from our `outside' point of view if Rob has a reasonable prior. But if Rob has a strange prior, this reasoning can look really bad. This is a problem because one of the core motivations for the Off-Switch Game is that it is very hard to specify a reward function which correctly captures our preferences. But, the idea goes, we don't have to worry about specifying the correct reward function because Rob will always defer to us. Presumably it is equally hard to specify the correct prior over our preferences. But in the presence of misleading signals, we \emph{do} have to worry about Rob's prior. Depending on its prior, Rob might or might not let itself be switched off. Thus, the Off-Switch Game does not successfully solve the problem it was designed to solve: making sure AI agents defer to us without having to worry about the details of their reward function and prior. Given that the assumptions of Good's theorem hold, AI agents still have an incentive to `listen to us' but, depending on their prior, they might decide to ignore our signals. This seems like cold comfort.

Now suppose the stakes are higher. Rob is contemplating a permanent change to our environment, perhaps a plan to stop climate change which, as side effect, permanently turns the sky orange \parencite[p.~202]{Russell2019}. Rob is quite confident that this is the right option and has some uncertainty about whether it will learn our actual preferences, so it goes ahead and implements this plan without asking. This seems like a situation where we really want Rob to defer to us. If Rob has an incentive to not defer, this is a problem.

\section{Practical Significance}

You might respond as follows. Grant that there is a possibility AI agents will not defer to us. But how likely is this possibility? It might be that AI agents defers to us in almost all cases. Should we still be worried about the tiny minority of cases where they fail to defer?\footnote{Thanks to an anonymous referee for raising this concern.}

There are two responses. First, even one instance of non-deference might have catastrophic consequences. AI agents might have powerful capacities to change our world in irreversible ways. This means that once the change is rolled out, it might be impossible to roll it back. One of the most worrying examples for an irreversible change is human extinction. But there are less drastic examples as well. AI agents might use up some non-renewable resource, permanently change our preferences or  turn the sky orange.

Second and more importantly, we don't have good reasons to think that the probability of non-deference is extremely low. Phenomena like sending misleading signals about one's preferences seem widespread. It seems implausible to ignore such phenomena when considering whether AI agents will defer to us in real life. So looks like there is not just an in-principle possibility but a substantial probability that AI agents will fail to defer. It is difficult to see why we should be confident that in practice, non-deference is extremely unlikely. Perhaps proponents of CIRL could provide  an argument for why this is true. This argument would need to show that, among other things, AI agents are very likely to maximize expected utility, be certain of updating by conditionalization and have perfect access to our preferences. Why is that? This is an important question which proponents of CIRL should address. 

\hypertarget{a-dilemma-for-provably-beneficial-ai}{%
\section{A Dilemma for Provably Beneficial AI}\label{a-dilemma-for-provably-beneficial-ai}}

I have argued that the result of \textcite{Hadfield2017} relies on substantive assumptions. Even if we make AI agents uncertain of our preferences, this does not guaranteed that they will always defer to us. This highlights a more general dilemma for provably beneficial AI. To prove that AI agents always defer to us (or are beneficial in some other sense), you have to make some decision-theoretic and epistemological assumptions. Either you make strong or weak assumptions.

Strong assumptions, such as expected utility maximization, certain conditionalization and perfect access to our preferences, will allow you to prove interesting guarantees. But such assumptions might not apply to all AI agents. As we have seen, there are reasons to think that AI agents in the real world might not be certain of updating by conditionalization and have perfect access to our preferences. So even if you can prove that given such assumptions, AI agents will be beneficial, this isn't much comfort if the assumptions might not be satisfied by many AI agents. Weak assumptions apply to a wider range of possible AI agents but don't allow you to prove much. As we have seen, if AI agents assign some positive probability to failures of conditionalization or to humans sending misleading signals about their preferences, they are not guaranteed to defer to us. 

Perhaps there is a way to successfully navigate this dilemma. We might find decision-theoretic assumptions weak enough to cover (almost) all plausible varieties of AI agents and strong enough to prove interesting guarantees---assumptions which are `just right'. But since we know so little about what AI agents might look like, it is not clear whether this will work out.

\hypertarget{conclusion}{%
\section{Conclusion}\label{conclusion}}

\textcite{Hadfield2017} propose a model for making sure that AI agents defer to us by making them uncertain about our preferences. I have argued that their result relies on strong decision-theoretic and epistemological assumptions: AI agent maximize expected utility, are certain of updating by conditionalization and have perfect access to our preferences. These assumptions limit the scope of the model since they might not be satisfied by AI agents in the real world.

Everything I have said here is compatible with the broad idea that we shouldn't tell AI agents to maximize a particular reward function but rather `teach them as we go along'. But we need more solid foundations for this idea. The problem of making sure AI agents defer to us is not yet solved.

\printbibliography

\end{document}